%% file: sigconf.tex
\PassOptionsToPackage{table}{xcolor}
\documentclass[sigconf]{acmart}
\AtBeginDocument{%
  }

\copyrightyear{2026}
\acmYear{2026}

\setcopyright{cc}
\setcctype{by}

\acmConference[MM '26]
{Proceedings of the 34th ACM International Conference on Multimedia}
{November 10--14, 2026}
{Rio de Janeiro, Brazil}

\acmBooktitle
{Proceedings of the 34th ACM International Conference on Multimedia (MM '26), November 10--14, 2026, Rio de Janeiro, Brazil}

\acmDOI{10.1145/3767308.3835047}

\acmISBN{979-8-4007-2213-4/2026/11}

\usepackage{multirow}
\usepackage{booktabs}
\usepackage{colortbl}
\usepackage{pifont}
\newcommand{\Checkmark}{\ding{51}}
\definecolor{deepgreen}{RGB}{0,120,70}
\definecolor{deepred}{RGB}{170,40,40}
\definecolor{grey}{RGB}{200, 200, 200}
\definecolor{subtleblue}{RGB}{235,245,255}
\definecolor{best}{RGB}{220, 245, 220}      % light green
\definecolor{second}{RGB}{255, 235, 200}    % light orange
\definecolor{bad}{RGB}{255, 220, 220}       % light red
\definecolor{pos}{RGB}{220, 245, 220}      % light green
\definecolor{neg}{RGB}{255, 220, 220}       % light red
\begin{document}

\title{SCOUT: Self-Checking and Recovery-Aware Tool-Thought Agents for Ultra-Long Egocentric Video Reasoning}

\author{Keyang Zhong}
\email{zhongky23@mail2.sysu.edu.cn}
\affiliation{
  \institution{Sun Yat-sen University}
  \city{Guangzhou}
  \country{China}
}
\affiliation{
  \institution{Shenzhen Loop Area Institute}
  \city{Shenzhen}
  \country{China}
}

\author{Kuo Wang}
\email{wangk229@mail2.sysu.edu.cn}
\affiliation{
  \institution{Sun Yat-sen University}
  \city{Guangzhou}
  \country{China}
}

\author{Peng Liu}
\email{angrybirdpeng@gmail.com}
\affiliation{
  \institution{Guangdong OPPO Mobile Telecommunications Corp., Ltd.}
  \department{OPPO AI Center}
  \city{Shenzhen}
  \country{China}
}

\author{Quanlong Zheng}
\email{zhengquanlong@oppo.com}
\affiliation{
  \institution{Guangdong OPPO Mobile Telecommunications Corp., Ltd.}
  \department{OPPO AI Center}
  \city{Shenzhen}
  \country{China}
}
\author{Junlin Xie}
\email{jun0wanan@163.com}
\affiliation{
  \institution{The Chinese University of Hong Kong, Shenzhen}
  \city{Shenzhen}
  \country{China}
}

\author{Zhijia Liang}
\email{liangzhj56@mail2.sysu.edu.cn}
\affiliation{
  \institution{Sun Yat-sen University}
  \city{Guangzhou}
  \country{China}
}

\author{Yanhao Zhang}
\email{zhangyanhao@oppo.com}
\affiliation{
  \institution{Guangdong OPPO Mobile Telecommunications Corp., Ltd.}
  \department{OPPO AI Center}
  \city{Shenzhen}
  \country{China}
}

\author{Guanbin Li}
\email{liguanbin@mail.sysu.edu.cn}
\correspondingauthor
\affiliation{
  \institution{Sun Yat-sen University}
  \city{Guangzhou}
  \country{China}
}
\affiliation{
  \institution{Shenzhen Loop Area Institute}
  \city{Shenzhen}
  \country{China}
}

\renewcommand{\shortauthors}{Zhong et al.}

\input{chapters/0abstract}

\begin{CCSXML}
<ccs2012>
   <concept>
       <concept_id>10010147.10010178.10010224.10010225.10010227</concept_id>
       <concept_desc>Computing methodologies~Scene understanding</concept_desc>
       <concept_significance>500</concept_significance>
       </concept>
   <concept>
       
       <concept_id>10010147.10010178.10010187.10010193</concept_id>
       <concept_desc>Computing methodologies~Temporal reasoning</concept_desc>
       <concept_significance>300</concept_significance>
       </concept>
   <concept>
       <concept_id>10010147.10010178.10010179.10003352</concept_id>
       <concept_desc>Computing methodologies~Information extraction</concept_desc>
       <concept_significance>100</concept_significance>
       </concept>
 </ccs2012>
\end{CCSXML}

\ccsdesc[500]{Computing methodologies~Scene understanding}
\ccsdesc[300]{Computing methodologies~Temporal reasoning}
\ccsdesc[100]{Computing methodologies~Information extraction}

\keywords{Ultra-long egocentric video, Multimodal, Chain-Of-Tool-Thought, Reinforcement Learning}


\maketitle

\input{chapters/1intro}
\input{chapters/2related_work}
\input{chapters/3method}
\input{chapters/4experiments}
\input{chapters/5conclusions}
\begin{acks}
This work is supported in part by the National Key R\&D Program of China (Nos.2024YFB3908503 and 2024YFB3908500), in part by the National Natural Science Foundation of China (No. 62322608) and in part by the Shenzhen Loop Area Institute under grant FPF10120260001. The authors would like to thank National Supercomputer Center in Guangzhou for providing high performance computational resources.
\end{acks}
% Keep Acknowledgments / References off the 8-page main-body budget.


\clearpage

% \section*{Ethics and Privacy Statement}

% This section of your ACM work should discuss the potential societal
% risks that might result from its publication; two to three sentences
% related to the findings of your study, or new advancements made
% possible by their developed methods. The privacy and ethics statement
% should clearly address the broader impacts of their work as it relates
% to the authors' interpretation of privacy, fairness, safety, human
% rights, data sovereignty, or future misuse and any benefit/risk
% trade-off resulting from this research. We acknowledge that some
% papers may have minimal societal risks beyond those considered by
% institutional review boards, and the dimensions considered by any
% review of the user study design or dataset licenses could be provided
% in this statement.

\bibliographystyle{ACM-Reference-Format}
\bibliography{sample-base}

\appendix

\end{document}

%% file: chapters/0abstract.tex
\begin{abstract}
Ultra-long egocentric video understanding requires reasoning over temporally sparse evidence distributed across hours or days, challenging current multimodal models with limited context and the grounding of key video segments.
While Chain-of-Tool-Thought (CoTT) agent systems enable iterative retrieval and inspection, they suffer from error propagation due to rigid zoom-in strategies that lack recovery mechanisms.
In this work, we address these challenges through SCOUT (Self-Checking Chain-Of-Tool-thought), a recovery-aware agentic framework introducing an adaptive policy that evaluates intermediate tool observations and dynamically trades off exploitation (zoom-in) and exploration (region switching), enabling robust multi-hop reasoning over extremely long horizons.
However, training such multi-turn tool-using agents remains challenging, as existing RL methods rely on sparse outcome-level rewards and lack supervision over extended decision trajectories, resulting in suboptimal credit assignment for long-horizon reasoning.
To address this, we develop UPS-GRPO, an uncertainty-prioritized policy optimization method that concentrates exploration on high-uncertainty post-tool states while preserving sample efficiency. We further introduce a turn-level advantage decomposition that integrates outcome rewards with tool-grounded temporal alignment rewards for improved credit assignment.
Experiments show that SCOUT achieves state-of-the-art results on ultra-long egocentric benchmarks, while remaining competitive on shorter-horizon long-video settings.

\end{abstract}

%% file: chapters/1intro.tex
\section{Introduction}
\begin{figure}[htbp]
    \centering
    \includegraphics[width=0.95\linewidth]{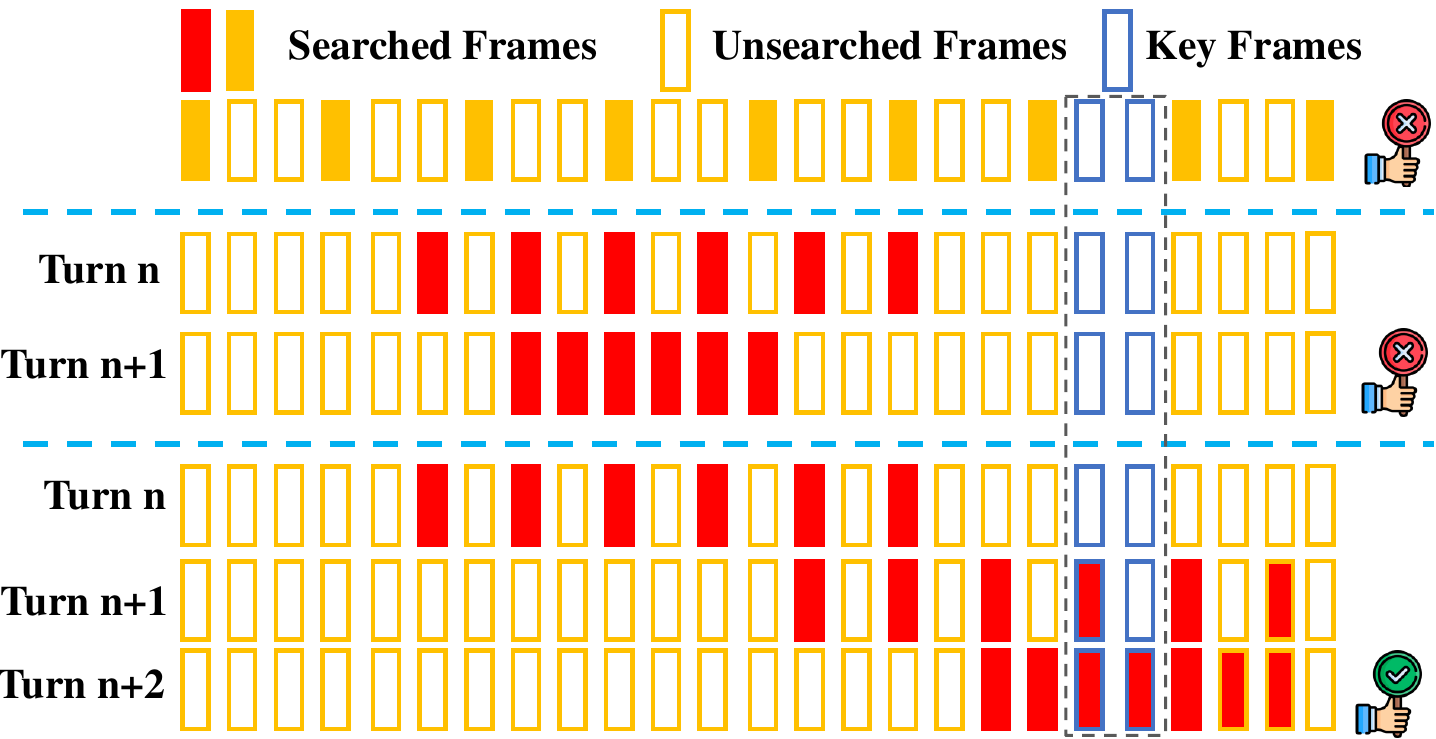} 
    \caption{Comparison of temporal search strategies for ultra-long video reasoning. Top: MLLM uniform sampling for initial visual observations. Middle: Failed search of existing agentic tool models due to irreversible early commitment and monotonic zoom-in. Bottom: Our SCOUT recovery-aware scheme, which dynamically switches strategies to locate answer key frames.}
    \label{fig1:scout_search_strategy}
\end{figure}

The proliferation of always-on egocentric recording devices has fueled growing interest in first-person AI assistants capable of drawing on a user's full visual history to deliver personalized support\cite{wu2024longvideobenchbenchmarklongcontextinterleaved}, alongside advances in person-centric visual recognition\cite{xu2026psgait,wu2026language,wu2025dagait}.  Unlike short-video question answering, ultra-long egocentric settings involve visual experience that unfolds continuously over hours or days, where relevant evidence is temporally sparse, distributed across long horizons, and often requires multi-hop aggregation across distant events\cite{grauman2022ego4dworld3000hours,yang2025egolifeegocentriclifeassistant,Chen_Di_Xie_2025}. These properties make it difficult for existing multimodal large language models (MLLMs) to operate effectively, as their reasoning is constrained by limited context windows and lossy visual compression, leading to a fundamental trade-off between temporal coverage and evidence fidelity\cite{shen2024longvu,shu2024videoxlextralongvisionlanguage}.

A promising way to break this trade-off is to view long-form video understanding as an \emph{agentic search problem} instead of a pure sequence modeling problem. Instead of processing the entire video in a single pass, recent Chain-of-Tool-Thought (CoTT) frameworks decompose reasoning into multiple steps, allowing a language model to call specialized tools for temporal retrieval, short-video inspection, and frame-level verification\cite{wang2024videoagent,egor12025,videothinker2026}. This hierarchical tool-use paradigm is especially appealing for long videos because it mirrors how humans search: first identify a plausible coarse time range, then zoom in to a smaller segment, and finally verify fine visual details.

However, existing CoTT-based agents typically adopt a rigid coarse-to-fine strategy that suffers from an \emph{irreversible early commitment} limitation\cite{ding2025videozoomer,pan2025timesearch,ye2025trethinkingtemporalsearch}. Once an initial temporal region is selected, reasoning typically refines within it, causing early localization errors to propagate without reassessing retrieved evidence or exploring alternatives~\cite{ye2025trethinkingtemporalsearch}. This kind of failure is particularly detrimental in ultra-long videos, where missing the correct temporal region early can permanently exclude relevant evidence from future reasoning. Thus, we propose SCOUT (Self-Checking Chain-Of-Tool-thought), a recovery-aware reasoning policy that augments CoTT with an explicit self-checking mechanism. SCOUT assesses the informativeness and consistency of tool observations, and dynamically selects between zoom-in refinement and temporal region switching, enabling robust multi-hop reasoning over extended horizons as shown in Figure~\ref{fig1:scout_search_strategy}. 

In multi-turn tool-use settings, the state distribution is shaped not only by model-generated tokens but also by external tool observations, leading to high uncertainty and distributional shift, especially at post-tool decision points\cite{arpo,yi2026pivotrlhighaccuracyagentic,jiang2025verltoolholisticagenticreinforcement,zhong2026rcgrporewardconditionedgrouprelative,ding2025empoweringmultiturntoolintegratedreasoning}. We formalize tool-augmented reasoning as a trajectory-level optimization problem and propose UPS-GRPO, an uncertainty-prioritized selection variant of GRPO. UPS-GRPO reallocates exploration budget toward post-tool states by selectively branching high-uncertainty continuations, while maintaining a constant effective group size for stable optimization and resource conservation. 

A trajectory with a wrong final answer may still contain useful retrieval and localization decisions, while a trajectory with a correct final answer may still include inefficient or spurious intermediate steps\cite{lightman2023letsverifystepstep,wei2025reinforcingmultiturnreasoningllm}. In ultra-long video understanding, this issue is even more pronounced because tool calls correspond to explicit temporal search decisions whose quality can often be assessed independently of the final answer\cite{ding2025empoweringmultiturntoolintegratedreasoning,wang2025sparlreinforcingllmagents,wu2026steppotentialadvantageestimation}. To this end, we introduce \emph{Turn-Based Tool-Use Advantage Semi-Decoupling}, which preserves final-answer correctness as the global optimization direction while rescaling trajectory-level advantages using tool-grounded turn-level signals such as semantic alignment and temporal overlap with clue intervals.

Effective training for long-form video agents requires data that teaches not only \emph{what} to answer, but also \emph{how} to search. We therefore construct a scalable data synthesis pipeline that generates high-quality CoTT trajectories with interleaved reasoning, tool calls, and observations, providing supervision for both supervised fine-tuning and reinforcement learning. To summarize, our contributions are as follows:
\begin{itemize}
    \item 
    We propose \emph{SCOUT}, a recovery-aware Chain-of-Tool-Thought reasoning framework for ultra-long video understanding that explicitly models search uncertainty through self-checking, enabling dynamic switching between local refinement and temporal re-exploration instead of irreversible monotonic zoom-in.
    \item 
    We introduce \emph{UPS-GRPO}, an uncertainty-prioritized reinforcement learning strategy for tool-augmented reasoning, together with a turn-level semi-decoupled advantage formulation that improves credit assignment by injecting tool-grounded intermediate signals while preserving trajectory-level optimization consistency.
    \item
    We establish an \emph{Ultra-Long Video CoTT} trajectory generation pipeline and show how it supports a two-stage SFT+RL recipe for training scalable long-video agents.

\end{itemize}

%% file: chapters/2related_work.tex
\section{Related Work}
\label{sec:related_work}

\subsection{Long Video Understanding}

Recent work on long-form and egocentric video understanding has moved from direct full-sequence encoding toward more scalable representations for hour-long or multi-day streams. Representative directions include compact visual compression, hierarchical memory, and structured temporal abstraction, such as compact long-video representations, event-centric memory, hierarchical agentic search, and temporal knowledge graphs\cite{learningcompactvideo,wang2025free,eventmemagenthierarchicaleventcentric,Liang_2026_CVPR,hierarchicallongvideo,sun2026egographtemporalknowledgegraph}. Benchmarks such as X-LeBench, HLV-1K, EgoLifeQA and HourVideo further show that success in ultra-long settings depends on locating sparse evidence across long temporal gaps rather than merely enlarging the context window\cite{xbench,zou2025hlv1klargescalehourlongvideo,yang2025egolifeegocentriclifeassistant,chandrasegaran2024hourvideo}. However, these methods mainly improve representation efficiency or memory capacity; they do not explicitly address the case where an early temporal hypothesis is wrong and the model must recover by revisiting alternative regions.

\subsection{Agentic Tool Use for Long-Video Reasoning}

A more closely related line of work formulates long-video understanding as an agentic search problem. VideoAgent first showed that an LLM can act as a controller over retrieval and inspection tools\cite{wang2024videoagent,xie2026streamrag}, and later systems strengthen this paradigm with hierarchical clip search, clue seeking, and frame-level verification\cite{li2025deepvideodiscovery,egor12025,videothinker2026,agenticverylong,lin2026videoseek,videoo3nativeinterleaved}. Recent variants also explore active navigation or hypothesis verification for long videos\cite{wang2025activevideoperceptioniterative,longvideor1smartnavigation,wang2026thinkverifyhypothesisverificationmultiagent,yang2025longvt,menon2025caviarcriticaugmentedvideoagentic,shi2026weaverendtoendagentictraining,yin2026videoarmagenticreasoninghierarchical,zhong2026collaborative}. Despite these advances, most agents still exhibit an implicit forward-only bias: once a candidate segment is selected, later steps mainly refine or verify within that neighborhood. As a result, they lack an explicit self-checking mechanism for deciding that the current evidence is weak or inconsistent and that the search should switch to a different temporal hypothesis.

\subsection{Reinforcement Learning and Training for Tool-Using Agents}

Training tool-using agents requires aligning model reasoning with external environment feedback. Furthermore, standard trajectory-level rewards often fail to provide granular credit assignment for intermediate search decisions that may be correct even if the final answer is flawed. ARPO shows that post-tool states are particularly uncertain and benefit from reallocating exploration to these decision points\cite{arpo}, while Tool-R1 and TSPO indicate that reward design and sampling policy optimization are important for stable multi-step tool use and long-video reasoning\cite{toolr1_2025,sun2025tspotemporalsampling}. On the data side, VideoThinker and ToolMind demonstrate that synthetic tool-interaction trajectories and turn-level filtering are effective for teaching agents how to search with tools\cite{videothinker2026,toolmind2025}. 

%% file: chapters/3method.tex
\begin{figure*}[htbp]
    \centering
    \includegraphics[width=0.95\linewidth]{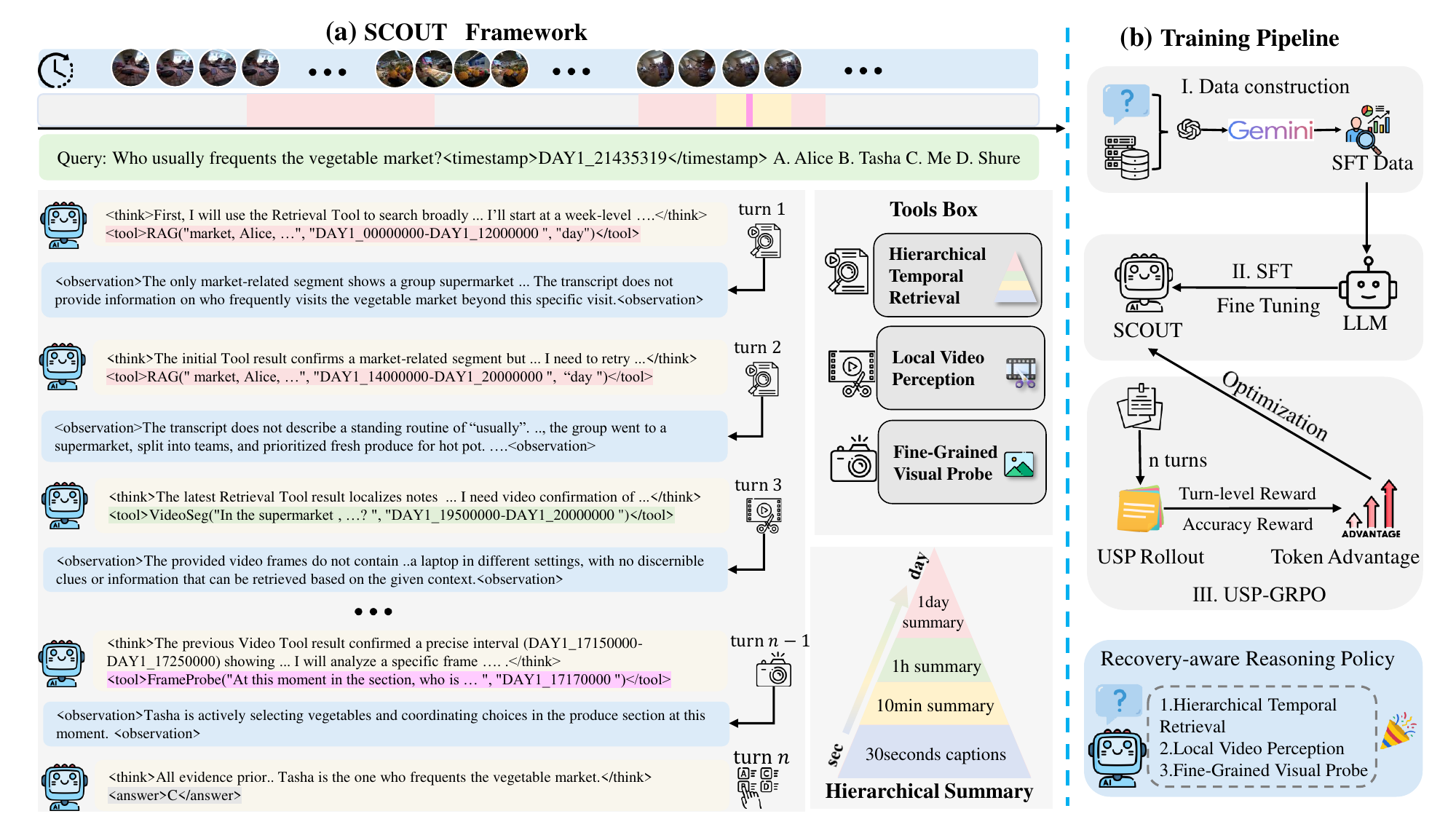} 
    \caption{(a) An illustration of the process of our proposed self-checking SCOUT. (b) Our three-stage training pipeline, consisting of data construction, Supervised Fine-Tuning (SFT), and Uncertainty-Prioritized Selection GRPO to optimize the policy.}
    \label{fig2:overview}
\end{figure*}

\section{Method}
\label{sec:method}

\begin{figure}[t]
    \centering
    \includegraphics[width=0.95\linewidth]{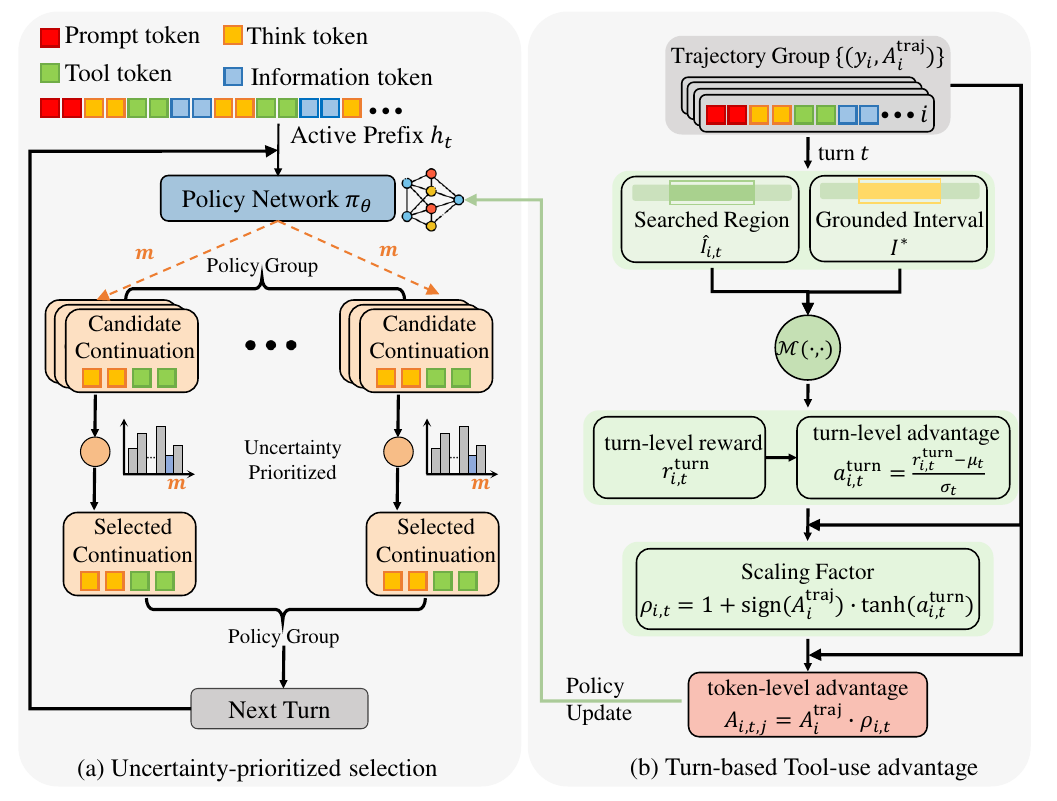}
    \caption{
    % UPS-GRPO: Uncertainty-Prioritized Selection and Turn-Based Tool-Use Advantage.
    % (a) Uncertainty-Prioritized Selection: sample multiple continuations at each state and retain only the highest-uncertainty branch for next turn rollout.
    % (b) Turn-Based Tool-Use Advantage: rescale trajectory-level advantages using normalized tool-grounded turn-level signals for improved credit assignment.
    UPS-GRPO. (a) Uncertainty-Prioritized Selection: sample multiple continuations and retain the most uncertain branch for the next rollout. (b) Turn-Based Tool-Use Advantage: rescale trajectory advantages with normalized turn-level tool signals for improved credit assignment.
    }
    \label{fig3:ups_grpo}
\end{figure}

As shown in Figure~\ref{fig2:overview}, our SCOUT framework does not process videos in a single passive pass; instead, it performs iterative reasoning loops to actively and strategically search for relevant information within the video before making decisions. When an initial search fails to identify the correct evidence region, subsequent steps can recover and refine the search, ultimately enabling accurate localization of the target clue range.

\subsection{Problem Formulation}

We study ultra-long video understanding as a multi-turn tool-augmented decision process. Given a video $V$ and a query $q$, the agent interacts with a toolbox over multiple turns to produce the final answer. At turn $t$, the agent observes the interaction history
\begin{equation}
    h_t = (q, y_1, o_1, \ldots, y_{t-1}, o_{t-1}),
\end{equation}
where $y_t={(z_t,a_t)}$ includes the thought process and tool-use action generated by the policy, and $o_t$ denotes the corresponding tool observation returned by the environment. The action space consists of three video-oriented tools:
\begin{equation}
    \mathcal{A} = \{\texttt{RAG}, \texttt{VideoSeg}, \texttt{FrameProbe}\}.
\end{equation}

The \texttt{RAG} tool performs text-based temporal retrieval with arguments of granularity level, temporal range, and keywords; \texttt{VideoSeg} analyzes a localized video segment with arguments of temporal range and question; and \texttt{FrameProbe} inspects a single timestamp (i.e., the frame) for fine-grained visual evidence. The agent alternates between reasoning and acting until it emits a final answer $\hat{y}$. Further details about tools are provided in the appendix.

\subsection{SCOUT}
\subsubsection{Self-Checking Chain-Of-Tool-thought}
Let $S_t$ denote the candidate temporal region maintained by the agent at turn $t$. Standard zoom-in CoTT implicitly enforces
\begin{equation}
    S_{t+1} \subseteq S_t,
\end{equation}
which assumes that each observation justifies further refinement of the current region. In long-form video understanding, however, early observations can be ambiguous, weakly relevant, or even misleading. We therefore replace monotonic refinement with a self-checking search policy that allows both \emph{refinement} and \emph{recovery}.

We define a unified transition set for the next-step search region:
\begin{equation}
\mathcal{T}(S_t) = \{ S' \mid S' \subseteq S_t \;\; \text{or} \;\; S' \sim S_t \},
\end{equation}
where $S' \subseteq S_t$ denotes zoom-in refinement, and $S' \sim S_t$ denotes $S'$ not a subset of $S_t$.
% camera ready add:
Self-checking is achieved through the aforementioned non-monotonic transformation space, and the policy learns from recovered perceptual trajectories. When tool observations are uninformative, inconsistent with the query, or temporally mismatched with candidate evidence, the strategy abandons the current region and switches to another temporal region.

\subsubsection{UPS-GRPO: Uncertainty-Prioritized Selection GRPO}

\begin{figure*}[t]
    \centering
    \includegraphics[width=0.95\linewidth]{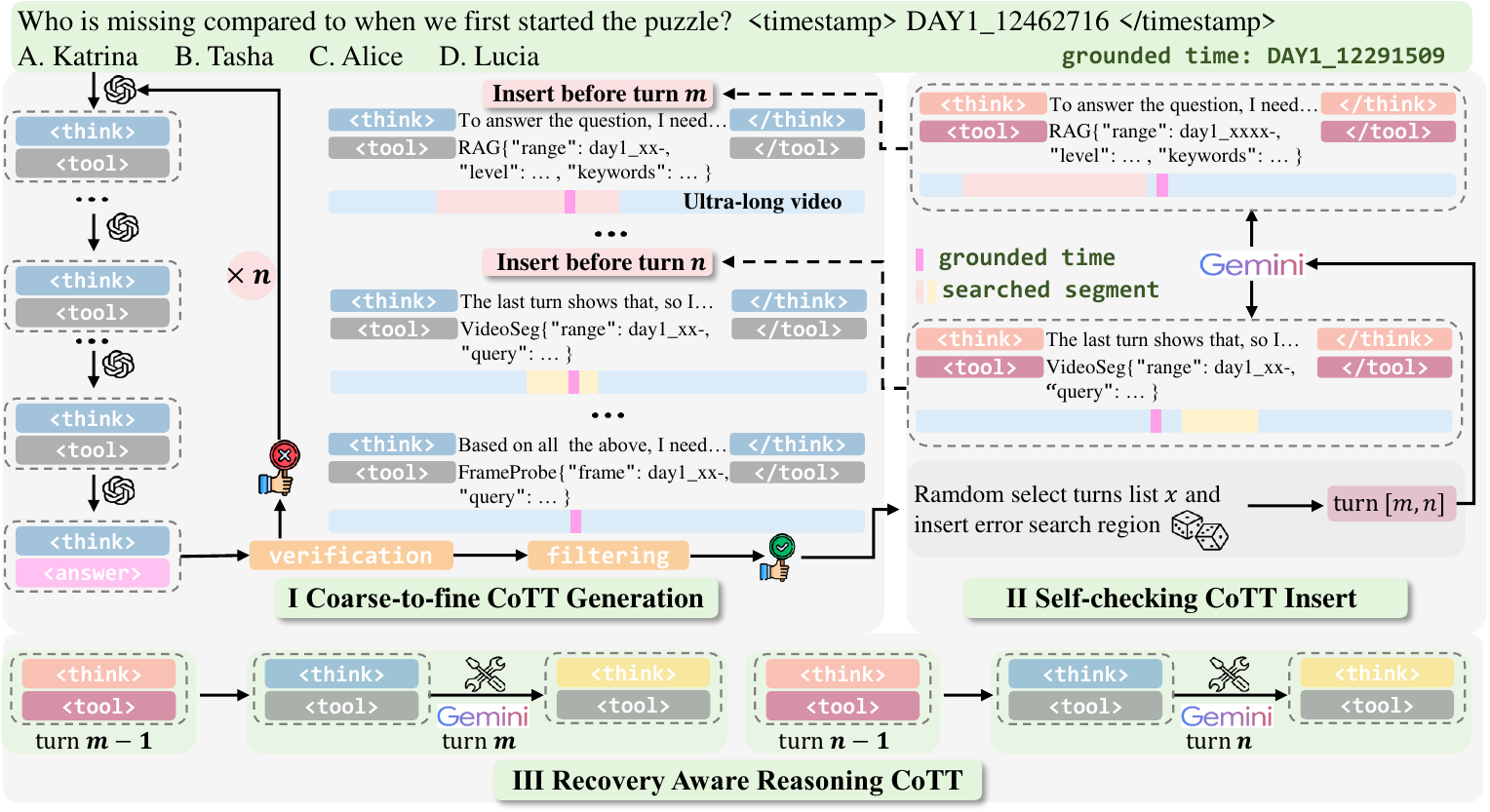}
    \caption{
    \textbf{Recovery-aware CoTT data construction pipeline.} Training trajectories are constructed through (1) constrained coarse-to-fine CoTT generation, (2) synthetic error injection with incorrect search segments, and (3) recovery-aware reasoning refinement for explicit self-correction, enabling robust multi-turn tool-use learning.
    }
    \label{fig4:data_construction}
\end{figure*}
We propose UPS-GRPO, a structured exploration and credit assignment framework for multi-turn tool-augmented reasoning. As illustrated in Figure~\ref{fig3:ups_grpo}, it consists of two key components: (a) uncertainty-prioritized selection and (b) turn-based tool-use advantage semi-decoupling.

\paragraph{Uncertainty-Prioritized Selection.}
In multi-turn tool-augmented reasoning, tool outputs are injected into the context as non-model-generated observations, inducing distributional shift and elevated uncertainty in subsequent inference. In ultra-long video, these post-tool states are decision-critical: the policy must determine whether to continue local refinement or to recover by exploring alternative regions. As shown in Figure~\ref{fig3:ups_grpo}(a), instead of uniformly allocating exploration budget across all trajectories, UPS-GRPO performs \emph{uncertainty-prioritized selection} at these high-uncertainty states. Given an active prefix $h_t$, we sample $m$ candidate continuations: 
\begin{equation}
    y_t^{(1)}, \ldots, y_t^{(m)} \sim \pi_{\theta}(\cdot \mid h_t)
\end{equation}
We estimate an uncertainty score $U(\cdot)$ for each candidate by using the average token-level log-likelihood, which reflects the model’s confidence over the generated sequence of this turn:

\begin{equation}
    U\!\left(y_t^{(k)}\right)
    = - \frac{1}{|y_t^{(k)}|} \sum_{j=1}^{|y_t^{(k)}|}
    \text{log}\pi_{\theta}(y_{t,j}^{(k)} \mid h_t, y_{t<j}^{(k)})
    \label{eq:ups_u}
\end{equation}

We then perform \emph{selection rather than expansion} by retaining only the highest-uncertainty continuation:
\begin{equation}
    k^\star = \arg\max_k U\!\left(y_t^{(k)}\right),
    \qquad
    \tilde{y}_t = y_t^{(k^\star)}.
    \label{eq:ups_select}
\end{equation}
The selected $\tilde{y}_t$ is used as the sole continuation for the next turn, while preserving the overall rollout budget.

This design concentrates exploration on trajectories where the policy is most uncertain, effectively reallocating capacity to decision-critical regions without increasing the effective group size. As a result, UPS-GRPO achieves a better exploration--efficiency trade-off for long-horizon tool-augmented reasoning.

% \paragraph{Turn-Based Tool-Use Advantage Semi-Decoupling}
\paragraph{Turn-Based Tool-Use Advantage.}
Final-answer rewards are highly sparse in long-horizon, multi-turn reasoning, as supervision is only available at the end of the trajectory. This makes it difficult to assign credit to intermediate decisions, even when they contain useful retrieval or localization steps. 
As illustrated in Figure~\ref{fig3:ups_grpo}(b), for each trajectory $i$ and turn $t$, we compute a turn-level reward based on temporal alignment:
\begin{equation}
    r_{i,t}^{\mathrm{turn}} = \max_{I^\star \in \mathcal{I}^\star} 
    \mathcal{M}\!\left(\hat{I}_{i,t}, I^\star\right),
\end{equation}
where $\hat{I}_{i,t}$ is the predicted search temporal region, $\mathcal{I}^\star$ is the set of ground-truth intervals, and $\mathcal{M}(\cdot,\cdot)$ unifies different temporal alignment measures, including segment IoU $|\hat{I}\cap I^\star|/|\hat{I}\cup I^\star|$ and point-wise grounding $\mathbf{1}[\tau \in I^\star]$. In addition, we define a trajectory-level reward $r_i^{\mathrm{traj}}$ based on the final answer correctness and format compliance. Within each GRPO group with size=$N$, we normalize turn-level rewards into turn-level advantages:

\begin{equation}
a_{i,t}^{\mathrm{turn}} = 
\frac{
r_{i,t}^{\mathrm{turn}} - 
\frac{1}{N}\sum_{i=1}^{N} r_{i,t}^{\mathrm{turn}}
}{
\sqrt{
\frac{1}{N}\sum_{i=1}^{N} \left(r_{i,t}^{\mathrm{turn}} - 
\frac{1}{N}\sum_{i=1}^{N} r_{i,t}^{\mathrm{turn}}
\right)^2
}
+ \epsilon
}.
\end{equation}

We then define a sign-preserving scaling factor:
\begin{equation}
    \rho_{i,t} = 1 + \mathrm{sign}(A_i^{\mathrm{traj}})\cdot\tanh\!\left(a_{i,t}^{\mathrm{turn}}\right),
\end{equation}
and assign token-level advantages as
\begin{equation}
    A_{i,t,j} = A_i^{\mathrm{traj}} \cdot \rho_{i,t}.
\end{equation}

Rather than directly injecting turn-level rewards $r_{i,t}^{\mathrm{turn}}$ into the trajectory objective (i.e., additive reward shaping), our formulation treats them as a \emph{multiplicative modulation signal} $\rho_{i,t}$ on the trajectory-level advantage $ A_i^{\mathrm{traj}}$. This design preserves the global optimization direction induced by the trajectory-level reward $r_i^{\mathrm{traj}}$, while allowing fine-grained, tool-grounded signals to softly adjust credit assignment across turns. As a result, it provides more stable training by avoiding conflicts between local and global objectives, while still improving temporal grounding sensitivity during intermediate reasoning steps.

\subsection{Data Construction}
We construct a recovery-aware reasoning Chain-of-Tool-Thought (CoTT) dataset through a three-stage pipeline as shown in Figure~\ref{fig4:data_construction}.

\paragraph{Stage I: Coarse-to-Fine CoTT Generation}
We begin by constructing reliable coarse-to-fine reasoning trajectories using GPT-4o\cite{achiam2023gpt} under a rigid decomposition strategy. To ensure high fidelity, each trajectory undergoes iterative sampling and correctness verification; specifically, if a generated answer is incorrect, the process is repeated up to five times, retaining only those paths that culminate in the correct final answer. The source data comprises question-answering pairs from Ego-R1 (training split)\cite{egor12025} and CG-Bench\cite{chen2024cg} (a subset with videos exceeding 30 minutes in duration). Because both datasets provide human-annotated grounded temporal intervals for necessary evidence, we rigorously filter the generated trajectories to ensure that no sequence exceeds seven tool-use turns and that every tool invocation temporally overlaps with the annotated ground truth. This initial phase yields a foundational set of approximately 2k high-quality, coarse-to-fine CoTT samples.

\paragraph{Stage II: Self-Checking CoTT Insert}
While the trajectories obtained in Stage I are correct, they lack exposure to failure modes that frequently arise in long-horizon reasoning, such as retrieving irrelevant temporal segments. To address this limitation, we utilize Gemini-2.5-Pro\cite{comanici2025gemini} to introduce controlled perturbations by injecting erroneous search segments into the obtained trajectories. Specifically, for each trajectory, we randomly select one or more turns and insert incorrect temporal ranges before those turns, simulating realistic retrieval mistakes. These inserted segments do not overlap with the grounded intervals and thus represent plausible but misleading evidence. Consequently, this augmentation expands the dataset to approximately 6k CoTT samples, successfully embedding common retrieval and reasoning mistakes while maintaining the overall structural integrity of the task.

\paragraph{Stage III: Recovery-Aware Reasoning CoTT}
The key limitation of Stage II is that, although errors are introduced, the subsequent reasoning steps do not explicitly acknowledge or correct these mistakes. In particular, for a pair of consecutive turns $(m-1, m)$, where turn $m-1$ corresponds to an inserted erroneous search segment and turn $m$ retrieves the correct evidence, the reasoning at turn $m$ typically proceeds as if no error had occurred. To address this, we further refine the trajectories by explicitly injecting recovery-aware reasoning. We employ Gemini-2.5-Pro\cite{comanici2025gemini} to systematically revise the reasoning at the specific recovery steps, explicitly bridging the gap when a current turn $m$ must recognize and correct an error from a previous turn $m-1$.  By explicitly integrating self-reflection and error-correction signals into the reasoning trace, the model demonstrates recovery-aware behavior, dynamically recognizing prior mistakes and adjusting its subsequent search strategy accordingly. 

Through this three-stage pipeline, we construct \textbf{RA-CoTT} (Recovery Aware Chain-of-Tool-Thought), a training dataset of $\sim$8k trajectories that jointly capture coarse-to-fine temporal grounding and recovery-aware reasoning. The pipeline progressively ensures correctness and grounding fidelity, injects realistic intermediate errors, and augments trajectories with explicit self-correction behaviors. RA-CoTT serves as training data for the first-stage supervised fine-tuning (SFT). Further details about the dataset and prompts are provided in the Appendix.

\subsection{Training Paradigm}

We adopt a SFT+RL training paradigm. First, supervised fine-tuning (SFT) is performed on synthesized \textbf{RA-CoTT} trajectories to initialize the policy with structured reasoning and tool-use behaviors. We then refine the policy via reinforcement learning under UPS-GRPO. Within this training pipeline, SCOUT improves the \emph{reasoning policy} via recovery-aware search, UPS-GRPO enhances \emph{exploration} through uncertainty-prioritized selection, and turn-based semi-decoupled advantages improve \emph{credit assignment}. Together, they enable stable and effective learning for long-horizon chain of tool thought augmented video reasoning under sparse supervision.

%% file: chapters/4experiments.tex
\section{Experiments}
\label{sec:experiments}
\input{tables/1main_table}
\input{tables/training_stages}

\begin{figure}[htbp]
  \centering
  \includegraphics[width=0.95\columnwidth]{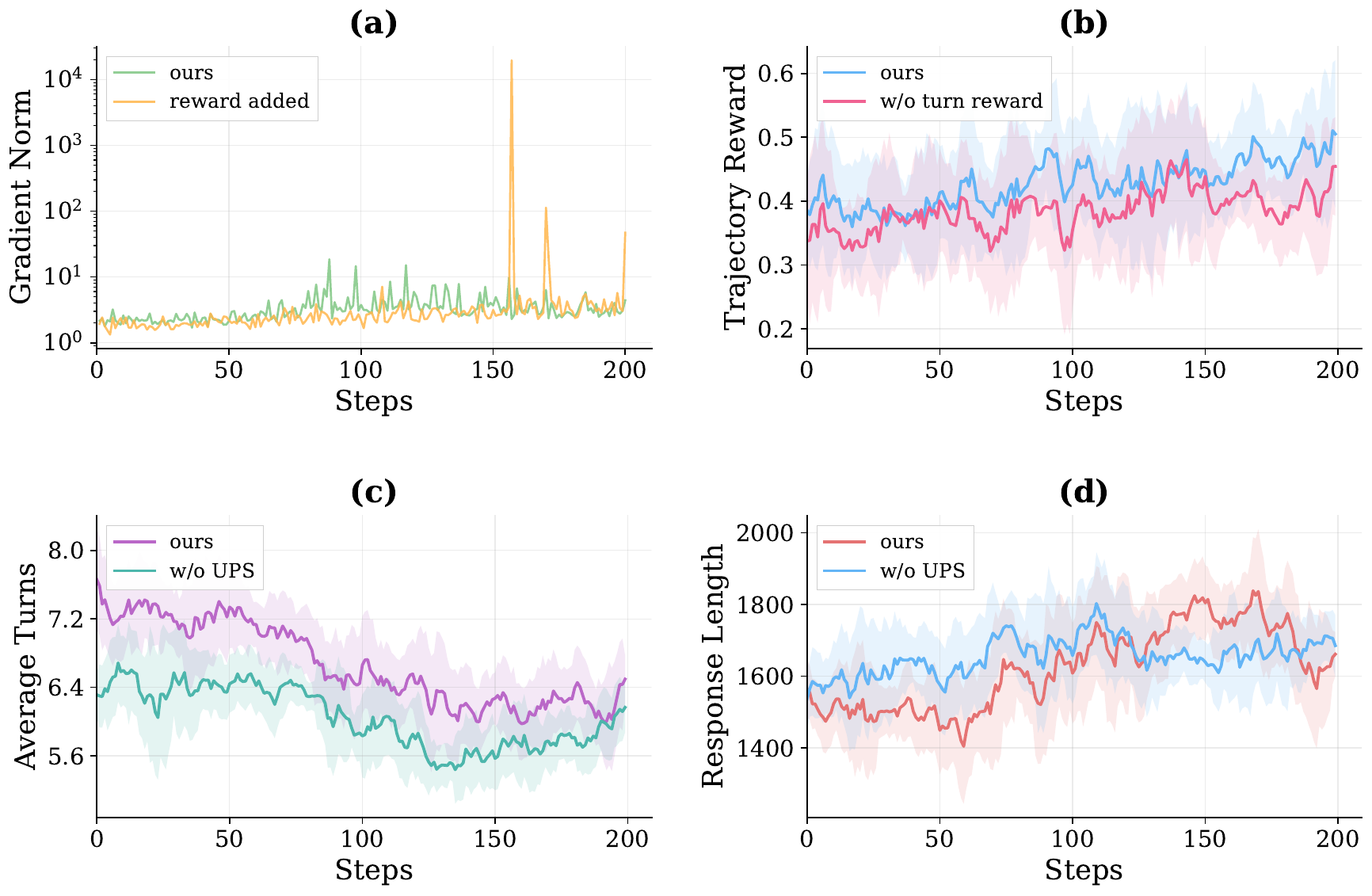}
  \caption{
    Training dynamics of our method and ablations.
    \textbf{(a)} Gradient norm;
    \textbf{(b)} Trajectory reward;
    \textbf{(c)} Average CoTT turns;
    \textbf{(d)} Mean response length (total words across turns).
  }
  \label{fig5:training-dynamics}
\end{figure}

\subsection{Experimental Setup}
\paragraph{Training setup.}
We initialize our model from Qwen2.5-7B-Instruct \cite{qwen2.5} due to its strong foundational capabilities and suitability. For cold-start initialization, we adopt the LLaMA-Factory framework\cite{zheng2024llamafactory}. Our reinforcement learning training is built upon verl\cite{sheng2024hybridflow}, which we extend to support multi-turn tool-calling scenarios.
Detailed information about the evaluation settings, implementation details, and more experimental results are provided in the appendix.

\paragraph{Baselines.}

We compare our method against three groups of baselines:
(1) \emph{Commercial multimodal models}, including strong closed-source APIs such as GPT-4o\cite{achiam2023gpt} and Gemini-1.5-Pro\cite{team2024gemini}.
(2) \emph{Open-source multimodal models}, including LongVU\cite{shen2024longvu}, Qwen3-VL\cite{bai2025qwen3}, LLaVA-Video\cite{llavavideo}, InternVL\cite{internvl3_5}, TSPO\cite{tspo2026}, Video-R1\cite{feng2025video}, and Rewatch-R1\cite{zhang2025rewatch}, which represent strong direct-video or long-context baselines.
(3) \emph{Agent-based systems} for long-video reasoning, including tool-using or retrieval-augmented approaches such as VideoAgent\cite{wang2024videoagent}, EgoGPT\cite{yang2025egolifeegocentriclifeassistant}, Ego-R1\cite{egor12025}, DVD\cite{li2025deepvideodiscovery}, TimeSearch-R\cite{pan2025timesearch}, LongVT\cite{yang2025longvt}, and Video-Zoomer\cite{ding2025videozoomer}.

\paragraph{Evaluation Benchmarks.}
We evaluate our method on four long-form video understanding benchmarks: \textbf{Video-MME(long)}\cite{fu2025video}, \textbf{EgoLifeQA}\cite{yang2025egolifeegocentriclifeassistant}, \textbf{Ego-R1 Bench}\cite{egor12025} and \textbf{HourVideo}\cite{chandrasegaran2024hourvideo}. These benchmarks cover complementary aspects of long-horizon multimodal reasoning, including temporal localization over long videos, egocentric memory-intensive question answering, and general long-video comprehension.

\textbf{Video-MME}\cite{fu2025video} evaluates long-video understanding under extended temporal ranges and comprises 900 videos, with 2700 MCQs. The benchmark is divided into Short, Medium, and Long subsets based on video length. We focus on the Long subset that consists of 300 videos that range from 30 to 60 minutes with 900 MCQs.

\textbf{EgoLifeQA}\cite{yang2025egolifeegocentriclifeassistant} consists of 500 long-context multiple-choice questions (MCQs) derived from the EgoLife dataset, where six participants cohabited for one week while continuously recording their daily activities. The benchmark focuses on approximately 50 hours of egocentric video from the perspective of Jake, one participant. The MCQs cover practical reasoning tasks such as object localization, event recall, habit tracking, and social interaction analysis. Additionally, every question is annotated with a query timestamp (e.g., 16:38 PM on Day 4) and a manually verified target timestamp, which identifies the specific temporal segment of the video containing the evidence required to answer the question correctly.

\textbf{Ego-R1 Bench}\cite{egor12025} comprises 300 question–answer (QA) pairs and evenly distributed across six first-person perspectives, same as EgoLifeQA with a query timestamp and a manually verified target timestamp. For each perspective, the benchmark includes a balanced mixture of human-annotated and human-verified QAs, ensuring both diversity and annotation reliability.

\textbf{HourVideo}\cite{chandrasegaran2024hourvideo} is a benchmark for hour-long video-language understanding. It consists of 500 egocentric videos (20–120 minutes) from Ego4D\cite{grauman2022ego4dworld3000hours}, paired with 12,976 five-way multiple-choice questions spanning diverse tasks, including summarization, perception, visual reasoning, and navigation.

\subsection{Main Results}

Table~\ref{tab:main_results} reports the main comparison on four long-form video understanding benchmarks spanning both exocentric and egocentric scenes. SCOUT-7B achieves the best results on the two ultra-long egocentric benchmarks. These gains are particularly notable because the two benchmarks require reasoning over 44.3 hours of continuous egocentric video, where answer evidence is temporally sparse, weakly signaled, and easily missed by policies that over-commit to an early hypothesis. We attribute this advantage primarily to the recovery-aware search policy introduced in SCOUT. In contrast to prior CoTT agents that largely follow a monotonic zoom-in trajectory once an initial temporal region is selected, SCOUT explicitly self-checks intermediate tool observations and can either refine the current region or recover by switching to a broader or neighboring region when the evidence remains ambiguous. This ability to revise the search trajectory is precisely what mitigates the irreversible early-commitment failure mode discussed in Section~\ref{sec:method}, and its benefit becomes more pronounced as the temporal horizon grows from tens of minutes to many hours.
Methods such as TimeSearch-R, LongVT and Video-Zoomer remain reasonably competitive on the shorter Video-MME(long) benchmark (average duration 41 minutes), yet deteriorate sharply on the egocentric benchmarks; for example, TimeSearch-R drops to 26.1\% on EgoLifeQA and 33.0\% on Ego-R1 Bench. DVD exhibits the same asymmetry even more clearly: it achieves 67.3\% on Video-MME(long), the best result among existing agent systems on that benchmark, but falls to 32.1\% and 31.0\% on EgoLifeQA and Ego-R1 Bench, respectively. Taken together, these results suggest that retrieval-heavy or strictly monotonic localization pipelines can be effective when the target evidence lies within a relatively short and well-structured search horizon, but they are far less reliable in ultra-long egocentric settings that require repeated hypothesis revision and multi-hop evidence aggregation. By contrast, SCOUT improves over the strongest prior agentic baseline Ego-R1 \cite{egor12025} by +9.1 and +6.0 points on EgoLifeQA and Ego-R1 Bench, while still maintaining a competitive 63.0\% on Video-MME(long) and 35.8\% on HourVideo. 

\subsection{Ablation Study}
\input{tables/mudules}
\input{tables/3ablation_ups}
\paragraph{SFT instills structure; RL refines decisions}
We ablate each training stage individually and in combination, finding that the full two-stage pipeline setting yields the best performance.
As shown in Table~\ref{tab:training_ablation}, removing SFT leaves the model with poor tool-use ability: it cannot reliably invoke tools ($\{\texttt{RAG}, \texttt{VideoSeg}, \texttt{FrameProbe}\}$). Consistently, the RL-only variant yields the lowest performance across all benchmarks and exhibits unstable behaviors during training, tending to follow superficial instructions and often misinterpreting returned information, rather than leveraging it as supporting evidence. 
SFT teaches the intended tool-use paradigm, inspecting tool outputs and incorporating the resulting information into the final answer. However, SFT remains imitation-driven\cite{NEURIPS2024_e0c9b65f}: it fits demonstrated formats, suffers from exposure bias, and fails to generalize under distribution shift. To address these limitations, we introduce RL with turn-level rewards, optimized via UPS-GRPO. This enables the policy to learn where to search within the video, what query-relevant information to extract, and how to recover from erroneous retrievals. As a result, RL pushes performance beyond the supervised ceiling on held-out videos, consistent with prior findings that GRPO enhances reasoning and generalization\cite{Guo_2025}.
Ultimately, this ablation demonstrates that while SFT provides the indispensable syntactic scaffold necessary to make complex tool-use computationally tractable, RL serves as the essential cognitive engine that elevates the model's grounded reasoning and temporal generalization well beyond the supervised ceiling.

\paragraph{Multiplicative Modulation vs. Additive Reward Shaping}
We compare our turn-based advantage modulation with a \emph{Turn Additive Reward} variant that directly injects turn-level rewards into the optimization objective. As shown in Table~\ref{tab:component_ablation}, the additive variant drops from \textbf{63.0} to 61.4 on Video-MME(long) and from 49.0 to 47.0 on Ego-R1 Bench. We attribute this to a fundamental difference in how intermediate supervision interacts with trajectory-level optimization: additive reward shaping alters the objective by introducing competing local signals at each turn, which can conflict with the final-answer reward and lead to misaligned credit assignment, especially in long-horizon settings. Our multiplicative modulation preserves the sign and direction of the trajectory-level advantage while adaptively reweighting its magnitude using tool-grounded signals, thereby achieving a form of \emph{direction-consistent credit refinement}. This design is further supported by the training dynamics in Figure~\ref{fig5:training-dynamics}(a), where our method exhibits significantly more stable gradient norms, indicating less noisy and more stable optimization. 

\paragraph{Efficiency Gains via Uncertainty-Prioritized Selection.}
As reported in Table~\ref{tab:component_ablation}, removing UPS leads to a noticeable performance drop on EgoLifeQA ($-1.8\%$) and Ego-R1 Bench ($-2.0\%$), despite a slight gain on the shorter-horizon VideoMME-Long. This discrepancy suggests that, in complex multi-turn tool-use scenarios, the ability to effectively handle high-uncertainty decision points is crucial for robust performance. More importantly, our longitudinal training analysis reveals an intriguing \emph{efficiency emergence} phenomenon. As shown in Figure~\ref{fig5:training-dynamics}(c), SCOUT-7B with UPS initially exhibits a higher average number of Chain-of-Tool-Thought (CoTT) turns than its ablated counterpart, reflecting more extensive exploration. However, as training progresses, it undergoes a more pronounced reduction in reasoning complexity. By the later stages of RL, the average CoTT turns converge to a level comparable to the baseline, while the final response length becomes significantly shorter. This behavior indicates that explicitly prioritizing exploration over high-uncertainty states enables UPS-GRPO to resolve ambiguous tool-interaction boundaries early in training, subsequently distilling these complex reasoning trajectories into more concise and efficient execution strategies.

We further conduct a controlled ablation by applying UPS-GRPO to the Ego-R1-SFT-3B\cite{egor12025} model, while keeping the training data and all other configurations fixed on the Qwen2.5-3B-Instruct base model\cite{qwen2.5} with Ego-R1. As shown in Table~\ref{tab3:ups_ablation}, UPS-GRPO consistently improves accuracy over standard GRPO, while simultaneously reducing the average number of reasoning turns. Unlike uniform expansion over all sampled continuations, UPS selectively prioritizes high-uncertainty trajectories, which typically correspond to decision-critical states induced by tool observations. This targeted exploration avoids allocating computational budget to low-uncertainty (already confident) branches, thereby reducing redundant refinement loops. Consequently, the policy learns to resolve ambiguity earlier and commit to more decisive actions, leading to shorter and more efficient reasoning chains.

\paragraph{The Efficacy of Turn-Level Credit Assignment. }
Removing the intermediate turn-level signal leads to a clear and systematic degradation in performance as shown in Table~\ref{tab:component_ablation}. This drop is not merely quantitative but indicative of a qualitative shift in the model’s learning dynamics: without intermediate guidance, the model exhibits less stable optimization and reduced effectiveness in long-horizon exploration, as reflected by consistently lower trajectory rewards throughout training as shown in Figure~\ref{fig5:training-dynamics}(b). This behavior suggests that relying solely on delayed, outcome-level supervision introduces severe credit assignment ambiguity. In long trajectories, the model lacks sufficient feedback to disentangle which intermediate decisions are causally responsible for success, leading to inefficient or even misleading policy updates. As a result, the learning process becomes both noisier and less sample-efficient, ultimately impairing downstream reasoning performance. In contrast, the presence of intermediate signals implicitly regularizes the exploration process by providing structured feedback during trajectory construction. This encourages the model to form more coherent and verifiable reasoning chains, rather than overfitting to spurious correlations between early actions and final outcomes. In long-horizon reasoning, effective optimization critically depends on introducing appropriately structured intermediate supervision to stabilize learning and guide credit assignment.

%% file: tables/1main_table.tex
\begin{table*}[htbp]
    % \small
    \footnotesize
    % \scriptsize
    \renewcommand{\arraystretch}{0.83}
    \centering
    \caption{Main comparison on four long video understanding benchmarks. We group baselines into three categories: Commercial Multimodal Models, Open-source Multimodal Models, and Agent-based Systems. All numbers are accuracy (\%). The best and second-best in each column is highlighted in \textbf{bold} and \underline{underline} respectively. $^{\dagger}$ indicates that our model outperforms all open-source models and is second only to commercial models. Results marked with $^{\ast}$ are reproduced by ourselves.}
    \label{tab:main_results}
    \resizebox{\textwidth}{!}{
    \begin{tabular}{lc|cc|cccc}
    \toprule
    
    \multirow{2}{*}{\textbf{Method}} & 
    \multirow{2}{*}{\textbf{Size}} &
    \multirow{2}{*}{\shortstack{\scriptsize\textbf{Reasoning}\\\scriptsize\textbf{Prompt}}} &
    \multirow{2}{*}{\scriptsize\shortstack{\textbf{Tool-Calling}\\\scriptsize\textbf{Self-Checking}}} &
    \multicolumn{1}{c}{\textbf{Exocentric}} & \multicolumn{3}{c}{\textbf{Egocentric}} \\
    \cmidrule(lr){5-5} \cmidrule(lr){6-8}
    & & & & \textbf{Video-MME (long)} & \textbf{EgoLifeQA} & \textbf{Ego-R1 Bench} & \textbf{HourVideo} \\

    \textit{\textbf{Average durations}} & & & & 41 min & 44.3 h & 44.3 h & 45.7 min\\
    \midrule
    
    \rowcolor{green!10}
    \multicolumn{8}{c}{\textit{Commercial Multimodal Models}} \\ \hline
    
    GPT-4o\cite{achiam2023gpt}                        & --      & -- & -- & 65.3 & 36.2 & 42.0 & \textbf{38.5} \\
    Gemini-1.5-Pro\cite{team2024gemini}               & --      & -- & -- & \textbf{67.4} & 36.9 & 38.3 & \underline{37.3} \\
    
    \midrule
    
    \rowcolor{green!10}
    \multicolumn{8}{c}{\textit{Open-source Multimodal Models}} \\ \hline
    Qwen3-VL-8B-Instruct\cite{bai2025qwen3}          & 8B      & \Checkmark & -- & 54.2 & 32.1 & 37.0 & 34.9 \\
    InternVL3.5-8B\cite{internvl3_5}                & 8B      & -- & -- & 51.4 & 33.9 & 32.0 & 33.8 \\
    LongVU\cite{shen2024longvu}                       & 7B      & -- & -- & 46.0 & 30.1 & 36.0 & 30.8 \\
    LLaVA-Video\cite{llavavideo}                     & 7B       & -- & -- & 52.7 & 36.4 & 29.0 & 34.6 \\
    TSPO\cite{tspo2026}               & 7B      & -- & -- & 56.4 & 30.7 & 38.0 & 35.4 \\
    Video-R1\cite{feng2025video}        & 7B      & \Checkmark & -- & 49.4 & 34.0 & 20.0 & 25.2 \\
    Rewatch-R1\cite{zhang2025rewatch}         & 7B      & \Checkmark & -- & 53.9 & 31.9 & 36.0 & 26.5 \\
    
    \midrule
    
    \rowcolor{green!10}
    \multicolumn{8}{c}{\textit{Agent-based Systems}} \\ \hline
    VideoAgent(gpt-4)\cite{wang2024videoagent}          & --    & -- & -- & 49.0 & 29.2 & 32.6 & -- \\
    DVD(o3+gpt4.1)\cite{li2025deepvideodiscovery}           & --    & \Checkmark & -- & \underline{67.3} & 32.1 & 31.0 & -- \\
    Ego-R1$^{\ast}$~\cite{egor12025}                       & 3B     & \Checkmark & -- & 53.0 & 38.5 & 43.0 & 35.6 \\
    EgoGPT\cite{yang2025egolifeegocentriclifeassistant}                        & 7B    & -- & -- & 48.9 & 36.0 & 41.0 & -- \\
    TimeSearch-R\cite{pan2025timesearch}        & 7B    & \Checkmark & -- & 56.0 & 26.1 & 33.0 & 22.8 \\
    LongVT\cite{yang2025longvt}             & 7B     & \Checkmark & -- & 62.6 & 25.3 & 29.0 & 33.0 \\
    Video-Zoomer\cite{ding2025videozoomer}        & 7B    & \Checkmark & -- & 55.8 & 25.7  & 29.0 & 33.9 \\
    
    \rowcolor{subtleblue}SCOUT-7B (RL-only)           & 7B     & \Checkmark & \Checkmark & 35.6 & 27.6 & 27.0 & 22.4 \\
    
    \rowcolor{subtleblue}SCOUT-7B (SFT-only)           & 7B     & \Checkmark & \Checkmark & 59.8 & \underline{44.0} & \underline{47.0} & 35.1 \\
    
    \rowcolor{subtleblue}\textbf{SCOUT-7B(Ours)}        & 7B    & \Checkmark & \Checkmark & 63.0 & \textbf{47.6} & \textbf{49.0} & 35.8$^{\dagger}$ \\
    
    \bottomrule
    \end{tabular}
    }
    \end{table*}

%% file: tables/training_stages.tex
\begin{table}[htbp]
\centering
\small
\setlength{\tabcolsep}{2.5pt}
\renewcommand{\arraystretch}{0.78}
\caption{
Ablation on training regimes.
All $\Delta$ values are computed with respect to the full model (SFT+RL).
}
\label{tab:training_ablation}
\begin{tabular}{cccccc}
\toprule
\multicolumn{2}{c}{Training Regime}
& Video-MME (long)
& EgoLifeQA 
& EgoR1 Bench 
& Format \\
\cmidrule(lr){1-2}
SFT & RL & Acc.\% & Acc.\% & Acc.\% & Acc.\% \\
\midrule

-- & \textcolor{deepgreen}{\Checkmark}
& 35.6 {\scriptsize \textcolor{deepred}{(-27.4$\downarrow$)}} 
& 27.6 {\scriptsize \textcolor{deepred}{(-20.0$\downarrow$)}} 
& 27.0 {\scriptsize \textcolor{deepred}{(-22.0$\downarrow$)}} 
& 50.6 {\scriptsize \textcolor{deepred}{(-49.4$\downarrow$)}} \\

\textcolor{deepgreen}{\Checkmark} & --
& 59.8 {\scriptsize \textcolor{deepred}{(-3.2$\downarrow$)}} 
& 44.0 {\scriptsize \textcolor{deepred}{(-3.6$\downarrow$)}} 
& 47.0 {\scriptsize \textcolor{deepred}{(-2.0$\downarrow$)}} 
& 99.8 {\scriptsize \textcolor{deepred}{(-0.2$\downarrow$)}} \\

\textcolor{deepgreen}{\Checkmark} & \textcolor{deepgreen}{\Checkmark}
& 63.0 
& 47.6 
& 49.0 
& 100.0 \\

\bottomrule
\end{tabular}
\end{table}

%% file: tables/mudules.tex
\begin{table}[htbp]
\centering
\small
\setlength{\tabcolsep}{2.5pt}
\renewcommand{\arraystretch}{0.78}
\caption{
Ablation on key components. Each module is removed from the full SCOUT-7B model. Parentheses denote the change ($\Delta$) relative to the full model.
}
\label{tab:component_ablation}
\begin{tabular}{lccc}
\toprule
\textbf{Method} 
& VideoMME-Long
& EgoLifeQA 
& EgoR1Bench \\
\midrule

\textbf{SCOUT-7B (Ours)}
& 63.0 & 47.6 & 49.0 \\

Turn Additive Reward
& 61.4 {\scriptsize \textcolor{deepred}{(-1.6$\downarrow$)}} 
& \textbf{48.0} {\scriptsize \textcolor{grey}{(+0.4$\uparrow$)}} 
& 47.0 {\scriptsize \textcolor{deepred}{(-2.0$\downarrow$)}} \\

w/o UPS 
& \textbf{63.4} {\scriptsize \textcolor{grey}{(+0.4$\uparrow$)}} 
& 45.8 {\scriptsize \textcolor{deepred}{(-1.8$\downarrow$)}} 
& 47.0 {\scriptsize \textcolor{deepred}{(-2.0$\downarrow$)}} \\

w/o turn reward 
& 61.2 {\scriptsize \textcolor{deepred}{(-1.8$\downarrow$)}} 
& 44.0 {\scriptsize \textcolor{deepred}{(-3.6$\downarrow$)}} 
& \textbf{50.0} {\scriptsize \textcolor{grey}{(+1.0$\uparrow$)}} \\

\bottomrule
\end{tabular}
\end{table}

%% file: tables/3ablation_ups.tex
\begin{table}[htbp]
\centering
\small
\setlength{\tabcolsep}{3pt}
\renewcommand{\arraystretch}{0.78}
\caption{
Ablation on UPS in GRPO.
We evaluate UPS-GRPO on top of Ego-R1-SFT-3B~\cite{egor12025} under identical training settings. $\Delta$ values are computed with respect to the Ego-R1~\cite{egor12025}.
}
\label{tab3:ups_ablation}
\begin{tabular}{lccc}
\toprule
Method 
& Video-MME (long)
& EgoLifeQA 
& Ego-R1-Bench \\
\midrule

Ego-R1$^{\ast}$~\cite{egor12025}
& 53.0 
& 38.5 
& 43.0 \\

\quad {\footnotesize CoTT Turns}
& {\footnotesize 6.69}
& {\footnotesize 7.14}
& {\footnotesize 7.74} \\

\midrule

+ UPS-GRPO
& 52.1 {\scriptsize \textcolor{grey}{(-0.9$\downarrow$)}}
& 39.4 {\scriptsize \textcolor{deepgreen}{(+0.9$\uparrow$)}}
& 48.0 {\scriptsize \textcolor{deepgreen}{(+5.0$\uparrow$)}} \\

\quad {\footnotesize CoTT Turns}
& {\footnotesize 5.41 {\textcolor{deepgreen}{(-1.28$\downarrow$)}}}
& {\footnotesize 5.61 {\textcolor{deepgreen}{(-1.53$\downarrow$)}}}
& {\footnotesize 4.60 {\textcolor{deepgreen}{(-3.14$\downarrow$)}}} \\

\bottomrule
\end{tabular}
\end{table}

%% file: chapters/5conclusions.tex
\section{Conclusions}
In this work, we present \textbf{SCOUT}, a recovery-aware agentic framework for ultra-long-horizon video reasoning that augments Chain-of-Tool-Thought with explicit self-checking temporal recovery, transforming long-video understanding from monotonic zoom-in search into adaptive evidence-seeking reasoning. We further introduce \textbf{UPS-GRPO}, an uncertainty-prioritized selection strategy that improves long-horizon tool learning by prioritizing exploration at uncertain post-tool states and strengthening turn-level credit assignment. Experiments across four long-form video benchmarks demonstrate strong and consistent gains, suggesting that effective ultra-long video reasoning depends not only on hierarchical tool use, but fundamentally on recovery-aware exploration and tool-grounded intermediate supervision.